\documentclass[conference]{IEEEtran}
\IEEEoverridecommandlockouts
\usepackage{cite}
\usepackage{amsmath,amssymb,amsfonts}
\usepackage{algorithmic}
\usepackage{graphicx}
\usepackage{textcomp}
\usepackage[ruled]{algorithm2e} 
\usepackage{caption}
\usepackage{subcaption}
\usepackage{color}
\usepackage{multirow}
\usepackage{bbm}

\def\BibTeX{{\rm B\kern-.05em{\sc i\kern-.025em b}\kern-.08em
    T\kern-.1667em\lower.7ex\hbox{E}\kern-.125emX}}
\begin{document}

\title{Time Series Deinterleaving of DNS Traffic}

\author{\IEEEauthorblockN{Amir Asiaee T.}
\IEEEauthorblockA{
\textit{Ohio State University}\\
Columbus, OH\\
asiaeetaheri.1@osu.edu}
\and
\IEEEauthorblockN{Hardik Goel}
\IEEEauthorblockA{
\textit{Microsoft Corporation}\\
Redmond, Washington\\
hagoel@microsoft.com}
\and
\IEEEauthorblockN{Shalini Ghosh, Vinod Yegneswaran}
\IEEEauthorblockA{
\textit{SRI International}\\
Menlo Park, CA\\
\{shalini,vinod\}@csl.sri.com}
\and
\IEEEauthorblockN{Arindam Banerjee}
\IEEEauthorblockA{
\textit{University of Minnesota}\\
Minneapolis, MN\\
banerjee@cs.umn.edu}
}
\newcommand{\norm}[2]{\|#1\|_{#2}}
\newcommand{\kl}{D_{\text{KL}}}
\newcommand{\vl}{\text{Vol}}

\newcommand{\beq}{\begin{equation}}
\newcommand{\eeq}{\end{equation}}

\newcommand{\htn}{\hat{\theta}_{\lambda_n}}
\newcommand{\hdn}{\hat{\Delta}_n}

\newcommand\B{\mathbb{B}}
\newcommand\E{\mathbf{E}}
\newcommand\N{\mathbb{N}}
\renewcommand\O{\mathbf{O}}
\renewcommand\P{\mathbf{P}}
\newcommand\Q{\mathbf{Q}}
\renewcommand\S{\mathbb{S}}
\newcommand\R{\mathbb{R}}
\newcommand\T{\mathbf{T}}
\newcommand\W{\mathbb{W}}
\newcommand\Y{\mathbb{Y}}

\newcommand\1{\mathbbm{1}}
\newcommand\0{\mathbbm{0}}

\newenvironment{packed_enum}{
	\begin{enumerate}
		\setlength{\itemsep}{1pt}
		\setlength{\parskip}{0pt}
		\setlength{\parsep}{0pt}
	}{\end{enumerate}}


\newcommand{\bd}{\boldsymbol}
\newcommand{\mf}{\mathbf}

\newcommand{\g}{\mathbf{g}}

\renewcommand{\a}{\mathbf{a}}
\renewcommand{\c}{\mathbf{c}}
\newcommand{\e}{\mathbf{e}}
\newcommand{\f}{\mathbf{f}}
\renewcommand{\r}{\mathbf{r}} 
\renewcommand{\u}{\mathbf{u}}
\renewcommand{\v}{\mathbf{v}}
\renewcommand{\o}{\mathbf{o}}
\newcommand{\s}{\mathbf{s}}
\newcommand{\w}{\mathbf{w}}
\newcommand{\x}{\mathbf{x}}
\newcommand{\y}{\mathbf{y}}
\newcommand{\z}{\mathbf{z}}
\newcommand{\p}{\mathbf{p}}
\newcommand{\q}{\mathbf{q}}

\newcommand\X{\mathbf{X}}
\newcommand\A{\mathbf{A}}
\newcommand\I{\mathbf{I}}

\newcommand{\cC}{{\cal C}}
\newcommand{\cH}{{\cal H}}
\newcommand{\cK}{{\cal K}}
\newcommand{\cL}{{\cal L}}
\newcommand{\cM}{{\cal M}}
\newcommand{\cN}{{\cal N}}
\newcommand{\cP}{{\cal P}}
\newcommand{\cQ}{{\cal Q}}
\newcommand{\cS}{{\cal S}}
\newcommand{\cT}{{\cal T}}
\newcommand{\cV}{{\cal V}}
\newcommand{\cX}{{\cal X}}
\newcommand{\cY}{{\cal Y}}
\newcommand{\cF}{{\cal F}}
\newcommand{\cZ}{{\cal Z}}


\newcommand{\bA}{\mathbf{A}}
\newcommand{\bB}{\mathbf{B}}
\newcommand{\bC}{\mathbf{C}}
\newcommand{\bD}{\mathbf{D}}
\newcommand{\bE}{\mathbf{E}}
\newcommand{\bI}{\mathbf{I}}
\newcommand{\bM}{\mathbf{M}}
\newcommand{\bP}{\mathbf{P}}
\newcommand{\bQ}{\mathbf{Q}}
\newcommand{\bR}{\mathbf{R}}
\newcommand{\bU}{\mathbf{U}}
\newcommand{\bV}{\mathbf{V}}
\newcommand{\bW}{\mathbf{W}}
\newcommand{\bX}{\mathbf{X}}
\newcommand{\bY}{\mathbf{Y}}
\newcommand{\bZ}{\mathbf{Z}}

\newcommand{\bz}{\mathbf{z}}
\newcommand{\hZ}{\hat{Z}}
\newcommand{\hbZ}{\hat{\bZ}}
\newcommand{\hz}{\hat{z}}

\newcommand{\hU}{\hat{U}}
\newcommand{\hV}{\hat{V}}
\newcommand{\hu}{\hat{u}}
\newcommand{\hv}{\hat{v}}
\newcommand{\htth}{\hat{\ttheta}}

\newcommand{\pr}{\mathbb{P}}
\newcommand{\ex}{\mathbb{E}}

\newcommand{\hX}{\hat{X}}
\newcommand{\hY}{\hat{Y}}
\newcommand{\hatx}{{\hat{x}}}
\newcommand{\haty}{{\hat{y}}}


\newcommand{\vertiii}[1]{{\left\vert\kern-0.25ex\left\vert\kern-0.25ex\left\vert #1
		\right\vert\kern-0.25ex\right\vert\kern-0.25ex\right\vert}}

\newcommand{\BNi}{\bB_N^{(i)}}
\newcommand{\xai}{\x_i^{(a)}}
\newcommand{\yai}{\bY_i^{(i)}}
\newcommand{\Xai}{\bX_i^{(a)}}
\newcommand{\Yai}{\bY_i^{(i)}}

\newcommand{\m}{\boldsymbol{\mu}}
\newcommand{\llambda}{\boldsymbol{\lambda}}
\newcommand{\xx}{\boldsymbol{\chi}}
\newcommand{\Th}{\boldsymbol{\theta}}
\newcommand{\Eta}{\boldsymbol{\eta}}
\newcommand{\oomega}{\boldsymbol{\omega}}
\newcommand{\ppi}{\boldsymbol{\pi}}
\newcommand{\pphi}{\boldsymbol{\phi}}
\newcommand{\ggamma}{\boldsymbol{\gamma}}
\newcommand{\bbeta}{\boldsymbol{\beta}}
\newcommand{\aalpha}{\boldsymbol{\alpha}}
\newcommand{\ttheta}{\boldsymbol{\theta}}
\newcommand{\eepsilon}{\boldsymbol{\epsilon}}
\newcommand{\no}[1]{\norm{#1}{}}
\newcommand{\myref}[1]{(\ref{#1})}
\newcommand{\del}[2]{\frac{\partial #1}{\partial #2}}

\newtheorem{prop}{Proposition}
\newtheorem{lemm}{Lemma}
\newtheorem{corr}{Corollary}
\newtheorem{clm}{Claim}
\newcommand{\proofsketch}{\noindent{\itshape Proof Sketch:}\hspace*{1em}}

\newcommand{\cx}{{\hat{x}}}
\newcommand{\cy}{{\hat{y}}}
\newcommand{\bp}{{\bar{p}}}
\newcommand{\eqn}[1]{(\ref{eq:#1})}
\newcommand {\commentout}[1] {}



\def\ints{{{\rm Z} \kern -.35em {\rm Z} }}  
\def\smallints{{{\rm Z} \kern -.3em {\rm Z} }}  
\def\pints{{{\rm I} \kern -.15em {\rm N} }}      
\newcommand{\reals}{\mathbb R}
\newcommand{\naturals}{\mathbb N}
\newcommand{\integers}{\mathbb Z}

\newcommand{\lapn}{\lessapprox_{\,n}}
\newcommand{\gapn}{\gtrapprox_{\,n}}
\newcommand{\apn}{\approx_n}

\newcommand{\RR}{\mathbb R}
\def\cplx{{{\rm I} \kern -.45em {\rm C} }}       
\def\l2{\rm {\mathcal L}^{2}(\reals)}            

\newcommand{\seqz}[2]{\mbox{$\{#1_{#2}\}$}_{#2 \in \smallints }}
\newcommand{\seqn}[2]{\mbox{$\{#1_{#2}\}$}_{#2 \in \pints }}
\newcommand{\abs}[1]{\left|#1\right|}

\newcommand{\nr}{\nonumber}
\newcommand{\be}{\begin{eqnarray}}
\newcommand{\ee}{\end{eqnarray}}
\newcommand{\bea}{\begin{eqnarray}}
\newcommand{\eea}{\end{eqnarray}}
\newcommand{\beaa}{\begin{eqnarray*}}
	\newcommand{\eeaa}{\end{eqnarray*}}
\newcommand{\bnad}{\begin{nad}}
	\newcommand{\enad}{\end{nad}}

\newcommand{\gb}{\beta_{\infty}}
\newcommand{\mgb}{\tilde{\beta}_{\infty}}
\newcommand{\gbb}{\beta_{2}}
\newcommand{\mgbb}{\tilde{\beta}_{2}}
\newcommand{\Jf}{J_{\infty}}
\newcommand{\mmJf}{\overline{J}_{\infty}}
\newcommand{\mJf}{\tilde{J}_{\infty}}
\newcommand{\Jff}{J_{2}}
\newcommand{\mJff}{\tilde{J}_{2}}
\newcommand{\Jc}{{\calJ_{\infty}}}
\newcommand{\mJc}{{\tilde{\calJ}_{\infty}}}
\newcommand{\Jcc}{{\calJ_{2}}}
\newcommand{\ptq}{P_{3 \overline{Q}}}
\newcommand{\pq}{P_{Q}}

\newcommand{\di}{{\,\mathrm{d}}}
\newcommand{\latop}[2]{\genfrac{}{}{0pt}{}{#1}{#2}}

\newcommand{\lip}{\langle}
\newcommand{\rip}{\rangle}
\newcommand{\uu}{\underline}
\newcommand{\oo}{\overline}
\newcommand{\La}{\Lambda}
\newcommand{\la}{\lambda}
\newcommand{\eps}{\varepsilon}
\newcommand{\veps}{\varepsilon}
\newcommand{\rmT}{{\rm T}}
\newcommand{\rmW}{{\rm W}}
\newcommand{\Ga}{\Gamma}

\newcommand{\sign}{{\mbox{\rm sign}}}
\newcommand{\ang}{{\mbox{\rm ang}}}
\newcommand{\dist}{{\mbox{\rm dist}}}

\newcommand{\nin}{\in\!\!\!\!\!/\,}

\newcommand{\calA}{{\cal A}}
\newcommand{\calB}{{\cal B}}
\newcommand{\calC}{{\cal C}}
\newcommand{\calD}{{\cal D}}
\newcommand{\tcD}{{\tilde{\cal D}}}
\newcommand{\calE}{{\cal E}}
\newcommand{\calF}{{\cal F}}
\newcommand{\calG}{{\cal G}}
\newcommand{\calH}{{\cal H}}
\newcommand{\calI}{{\cal I}}
\newcommand{\calJ}{{\cal J}}
\newcommand{\calK}{{\cal K}}
\newcommand{\calL}{{\cal L}}
\newcommand{\calM}{{\cal M}}
\newcommand{\calN}{{\cal N}}
\newcommand{\calO}{{\cal O}}
\newcommand{\calP}{{\cal P}}
\newcommand{\calQ}{{\cal Q}}
\newcommand{\calR}{{\cal R}}
\newcommand{\calS}{{\cal S}}
\newcommand{\calT}{{\cal T}}
\newcommand{\calU}{{\cal U}}
\newcommand{\calV}{{\cal V}}
\newcommand{\calW}{{\cal W}}
\newcommand{\calX}{{\cal X}}
\newcommand{\calY}{{\cal Y}}
\newcommand{\calZ}{{\cal Z}}

\newcommand{\RE}{{\cal R}e}

\newcommand{\Prob}{{\rm Prob\,}}
\newcommand{\diam}{{\rm diam\,}}
\renewcommand{\mod}{{\rm mod\,}}
\newcommand{\sinc}{{\rm sinc\,}}
\newcommand{\ctg}{{\rm ctg\,}}
\newcommand{\ifff}{\mbox{\ if and only if\ }}
\renewcommand{\overline}{\bar}
\renewcommand{\widetilde}{\tilde}
\renewcommand{\widehat}{\hat}

\newcommand{\boldx}{{\bf x}}
\newcommand{\boldX}{{\bf X}}
\newcommand{\boldy}{{\bf y}}
\newcommand{\indic}{\mathbbm{1}}
\newcommand{\uux}{\uu{x}}
\newcommand{\uuY}{\uu{Y}}

\newcommand{\opt}{\rm{opt}}

\newcommand{\ffrac}[2]
{\left( \frac{#1}{#2} \right)}

\newcommand{\one}{\frac{1}{n}\:}
\newcommand{\half}{\frac{1}{2}\:}

\newcommand{\BAMS}{{\em Bulletin of the Amer. Math. Soc.}}
\newcommand{\TAMS}{{\em Transactions of the Amer. Math. Soc.}}
\newcommand{\PAMS}{{\em Proceedings of the Amer. Math. Soc.}}
\newcommand{\JAMS}{{\em Journal of the Amer. Math. Soc.}}
\newcommand{\LNM}{{\em Lect. Notes in Math.}}
\newcommand{\LNCS}{{\em Lect. Notes in Comp. Sci.}}
\newcommand{\IV}{{\em Invent. Math.}}
\newcommand{\JAM}{{\em J. Anal. Math.}}
\newcommand{\Sc}{{\em Science}}

\maketitle

\begin{abstract}

Stream deinterleaving is an important problem with various
applications in the cybersecurity domain.  In this paper, we consider
the specific problem of deinterleaving DNS data streams using
machine-learning techniques, with the objective of automating the
extraction of malware domain sequences.  We first develop a generative model for
user request generation and DNS stream interleaving.  Based on these
we evaluate various inference strategies for deinterleaving including
augmented HMMs and LSTMs on synthetic datasets.  Our results
demonstrate that state-of-the-art LSTMs outperform more traditional
augmented HMMs in this application domain.
\end{abstract}


		\section{Introduction}
	Deinterleaving temporal data streams is a general machine-learning
	problem with important applications to security and privacy.
	Specifically, interleaved network data streams are a common occurrence
	in cyber-threat monitoring which complicates many analyses.  In many
	instances, the individual stream identifiers are unavailable due
	to technical challenges, such as the vantage point of the data
	collector or are intentionally supressed to protect the privacy
	of users in the network.
	
	For example, consider packet traces collected in a local area network
	where the source IP addresses are removed, or data collected from the
	external-facing interface of a proxy server, or a NAT firewall
	where individual client identifiers are unavailable.  Detecting anomalous
	behavior, especially stealthy and low-volume attack patterns, in
	these aggregated noisy streams is significantly more challenging than
	in a traditional deinterleaved setting.
	
	In this paper, we discuss a variant of this problem, i.e.,
        deinterleaving client request streams from recursive DNS
        resolvers to mine threat intelligence.  Such DNS data streams
        are shared among Internet service providers (ISPs) through
        mediums such as the Security Information Exchange (SIE)
        ~\cite{gao2013empirical} and are a valuable source of
        intelligence to the cybersecurity community.  Here, the
        individual client requests to the recursive DNS resolver are
        typically suppressed and what we have are inter-resolver
        communications (i.e., communications between the recursive
        resolver and the root server, TLD servers and other secondary
        resolvers).  We are interested in the application of advanced
        machine-learning techniques to automate the extraction of
        malware domain groups~\cite{gao2013empirical} from such
        resolver streams.
	
	Malware infections while browsing the Internet have become very
	prevalent and occur due to various reasons such as drive-by exploits,
	phishing attacks etc~\cite{sophos, symantec}.  In a typical infection, the user starts from
	a landing page and then goes through a sequence of seemingly harmless
	intermediate websites, until reaching a site that contains the malicious
	exploit that harm the user by installing malware or stealing private
	data.  The intermediate sites are typically redirection chains implemented
	in JavaScript for the purpose of obfuscation.
	Even though many landing and exploit websites are continously identified and
	blacklisted, thousands of new malicious domains emerge daily.  However,
	pieces of the redirection infrastructure get reused across campaigns
	and thus the actual sequence of websites traversed by the user contains
	information that may help in quickly identifying new exploit sites.
	
	When a user makes a browser request to visit a website, it first
	resolves the domain name by asking its recursive resolver.  If the
	answer for the query is cached by the resolver the answer is
	immediately provided to the client. Otherwise, it initiates a set of
	recursive queries, leading to the final queried website's IP address.
	Each webpage may have several embedded objects from many domains
	leading to a sequence of domain lookup requests emanating from the
	client.  Tracking the set of DNS requests made by each client is thus
	a useful means to identifying new and emergent malware infection
	sequences.  However, to protect user privacy ISPs typically only
	capture data from the external facing interface of the recursive
	resolver, effectively suppressing the individual client stream
	identifiers.  As there are hundreds of users making requests at the
	same period of time, and all of these requests are pushed to a single
	queue of a local DNS resolver, we cannot tell apart individual user's
	sequences of requests and perfectly deinterleaving all requests for
	deanonymization purposes is impossible.  However, our objective is not
	deanonymization, but rather extraction of malware domain sequences
	which are observed repeatedly across resolvers.  We believe that advanced
	machine learning strategies could be  in such selective
	deinterleaving of DNS time-series for the extraction of malware
	domain groups.

	{\bf Prior Work. } To the best of our knowledge deinterleaving has not
	been applied to DNS resolver queue's data.  Some earlier
	work \cite{gao2013empirical} investigates the use of a sliding window
	approach to identify new malicious domains by exploring the domains
	that typically form neighbors of known malicious domains in the
	resolver queue, while ignoring the actual sequential information.  The
	challenges of applying existing deinterleaving methods to DNS data is
	twofold.  First, most of the methods has been designed for
	deinterleaving Markov chains \cite{batu2004inferring,
		seroussi2009deinterleaving, seroussi2012deinterleaving,
		minot2014separation} and HMMs \cite{landwehr2008modeling}, and as we
	will discuss in Section~\ref{sec:gen}, the dynamics of submitting new queries to
	the local resolver is more complicated than simple Markov chain or
	HMM.  Moreover, the state space of the models and number of sequence
	sources are very small in previous work
	applications \cite{minot2014separation, landwehr2008modeling}, while
	in our application, huge number of websites explodes the size of state
	space and also tens of users may be active in a network
	simultaneously.  Because of the nature of our dataset, we need to use
	tools other than those adopted in
	literature \cite{minot2014separation,
		landwehr2008modeling,burge1998finding, burge1997prediction}.
	
	Another very useful model for time-series is Recurrent Neural Networks
	(RNN).  Recently, RNNs and their variants (Gated Recurrent Units
	(GRUs)\cite{chung2014empirical}, Long Short-Term Memory
	(LSTMs) \cite{Hochreiter}) have seen a lot of success in modeling
	time-series in multiple domains \cite{bahdanau2014neural,
		NIPS2008_3449, sutskever2014sequence}.  However, to our knowledge even
	simple RNN tools have not been applied to the deinterleaving
	problem. Using RNN-type tools for deinterleaving mixed DNS request
	logs is a completely unexplored area.  Motivated by the power of LSTMs
	to model non-linear dependencies, we seek to apply LSTMs to such data
	and start a new direction of work towards identifying newer malicious
	domains more efficiently.
	
{\bf
        Contributions. } This paper presents a preliminary exploration
          of the utility of various machine-learning models to address
          the time series deinterleaving problem for malware domain
          group extraction.  Specifically, we present a model for DNS
          request generation and resolver-sequence interleaving and
          evaluate the utility of various inference strategies on
          sythetic examples including Augmented Hidden Markov Models
          (AHMMs) and LSTMs finding that LSTMs outperform AHMMs.  
          Extending this analysis to real and
          large-scale datasets is future work.
	
		\section{Problem Formulation}
	\label{sec:gen}
        \label{subsec:nature}

	

	A user starts by visiting a page e.g., {\tt a.com}. While
        launching the webpage, many queries are being generated from
        different components of that browsed webpage: {\tt a.com, ad1.com,
        audio1.org}.  In another scenario a webpage can redirect the user to a sequence of other pages and generate sequence of requests.     
    We refer to this sequence as a {\em query episode}.
        Next, when the user opens a new website another episode 
        is started.  The same process generates query
        sequences for other users.  For example, a second user generates: 
        {\tt b.com, ad2.com} and after the interleaving we may observe the
        following sequence in the resolver:{\tt  b.com, a.com, ad1.com,
        ad2.com, audio1.org}.
	We call this process \emph{request interleaving}.
	Our goal is to deinterleave the two request sequences. 
	
	\subsection{User's Request Generation Model}
	\label{subsec:user}
	The \emph{browsing process} of a user can be modeled as simple
        as a Markov chain (MC) of webpages, an HMM, or an HsMM model.
        Figure \ref{fig:hsmm} illustrates these three different user
        model.	
        We model the browsing process described above
        using a Hidden Semi-Markov Model (HsMM).  MC and HMM are
        special cases of this process.  The hidden layer of the HsMM
        consists of random variables $W$ representing the browsed
        webpages.  Note that pages are hidden because what we see are
        only the DNS requests.

	
	The page transition matrix is different for each user and is
        represented by the matrix $\P_u$.  
        The observed state of the HsMM is the domain name request $R$
        which will be put in the resolver queue.
	Note that the time between subsequent browsed pages (which is equal to the time spend in a page before moving to the next one) in reality is different from the duration parameter in our model.
	In real world data, each user spends an interval on a page but in our model since we are only interested in the order of queries, we only count the number of requests that the page will query from the resolver and represent it by the random variable $D$.
	So the duration parameter $D$ represents the number of outstanding requests from the current page.
	
	
	Fig \ref{fig:hsmm2} shows the details of the model and Table \ref{tab:params} summarizes the model parameters.
	$\O_u(w, r)$  is the probability of submitting (outputting/observing) request $r$ on the webpage $w$, for the user $u$.	
	Conditional probabilities of the model are as follows:	
	\begin{figure}
	\centering
	\subcaptionbox{
		Markov Chain
	}{\includegraphics[width=0.11 \textwidth,]{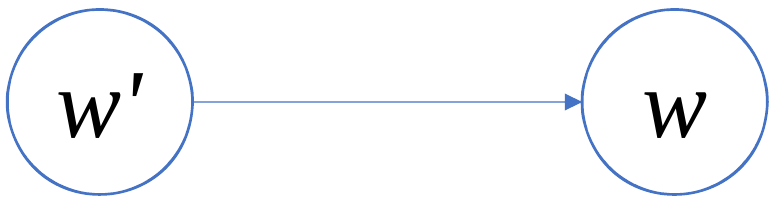}}~
	\subcaptionbox{
		HMM
	}{\includegraphics[width=0.11 \textwidth,]{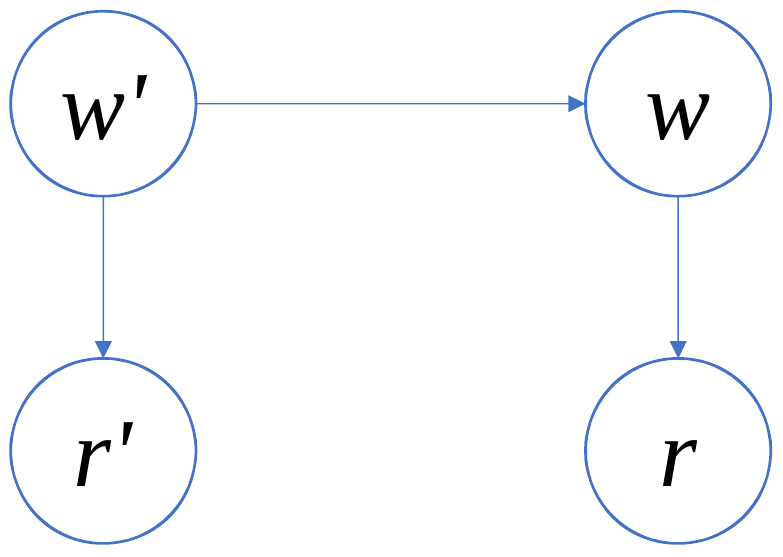}}~
	\subcaptionbox{
		HsMM
		\label{fig:hsmm2}
	}{\includegraphics[width=0.13 \textwidth]{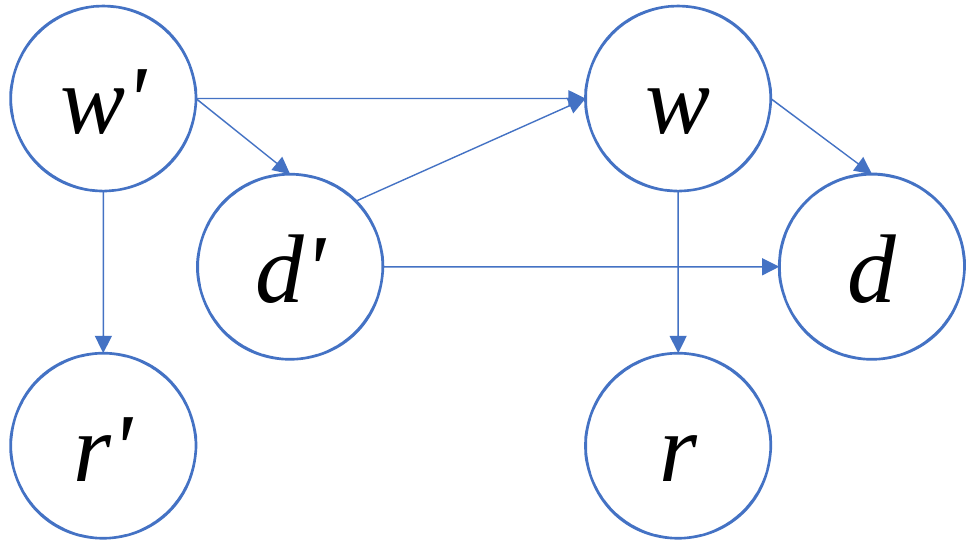}}
	\caption{User's browsing models.}
	\label{fig:hsmm}
\end{figure}	
	\begin{equation}
	\label{eq:probs}
	\begin{alignedat}{3}
	\pr_u(w | w', d') &= 
	\begin{cases} 
	[\bP_u]_{w'w} & d' = 1  \\
	\delta(w, w') & d' > 1
	\end{cases} &,&
	\\ 
	\pr_u(d | w, d') &= 
	\begin{cases} 
	p_{w}(d) & d' = 1 \\
	\delta(d, d'-1) & d' > 1
	\end{cases} &,&
	\\   
	\pr_u(r|w) &= [\O_{u}]_{w, r}&,&	
	\end{alignedat} 
	\end{equation}	
	The duration parameter cannot be zero, when $d = 1$ (i.e., the page's last request is submitted) and the user moves to another page which resets the duration using $p_w(d)$. 
	The duration probability $p_w(d)$ determines the number of requests that the webpage $w$ will query and is independent of user $u$. 
	
	\begin{table}
		\centering
		\begin{tabular}{|c|l|}
			\hline
			Symbol & Explanation \\ 
			\hline 
			$W$ & RV for the webpage \\ \hline
			$D$ & RV for the number of requests to be issued on a page \\ \hline
			$R$ & RV for the issued DNS request \\ \hline
			\hline 
			$m$ & Number of users \\ \hline
			$n$ & Total number of pages \\ \hline
			$q$ & Maximum number of requests per page \\ \hline 
			\hline 
			$\P_u \in \reals^{n \times n}$ & Webpage transition matrix of the user $u$ \\ \hline 
			$p_w(d), d \in [q]$ & Distribution of number of requests $d$ on the page $w$ \\ \hline 
			$\O_u \in \reals^{n \times n}$ & Output distribution matrix of the user $u$  \\ \hline
		\end{tabular}
		\caption{Summary of the model parameters and random variables (RV). For each random variable the corresponding small letter represents a realization. Note that $W$ and $D$ depend on the user but to avoid cluttering we omitted the index $u$.}
		\label{tab:params}
	\end{table}

	\subsection{Resolver's Sequence Interleaving Model}
	\label{subsec:interle}
	Each time step in our model is a slot in the resolver's queue. 
	Since there cannot be two requests in the same slot, only one user out of $m$ can fill the $t$-th slot of the queue. 
	Considering the frequency of request generation, we assume that each user $i$ has a probability of $\alpha_i$ to generate the $t$-th request where $\sum_{i=1}^{m} \alpha_i =1$. 
	So if a user is very active it has higher $\alpha_i$ and submit requests more often. 
	
	In a more complicated setting, one can model the the ``turn'' of $m$ users as a Markov chain.
	We name the transition matrix of the user's Markov chain as $\A = [\alpha_{ij}] \in \reals^{m \times m}$.
	Therefore the probability of user $j$ generating the $t$-th request from the $i$-th user is $\pr(U(t)=j|U(t-1)=i) = \alpha_{ij}$.
	The random variable $U(t) \in [m]$ represents the active user that generated the $t$-th request of the resolver queue. 
	As mentioned above, a simplified variant of the \emph{user's transition matrix}  $\A$ is the \emph{shares vector} $\aalpha$ that has been used in literature \cite{minot2014separation, batu2004inferring} where $\forall i, j: \pr(U(t) = i|U(t-1) = j) = \alpha_i$.
	

	To distinguish each user's corresponding HsMM random variable in the interleaving process we use both user index and time index. For example, $W_{k}(t)$ is the user $k$'s current webpage. 
	Note that here the time is different from the real world time and HsMM duration that discussed in Section \ref{subsec:user}. 
	Time here is just an index into the resolver's sequence of queries.
	For example, $W_{k}(t)$ shows the webpage of user $k$ when the $t$th request was submitted to the resolver. 
	
\begin{figure}
		\centering
		\includegraphics[scale=.4]{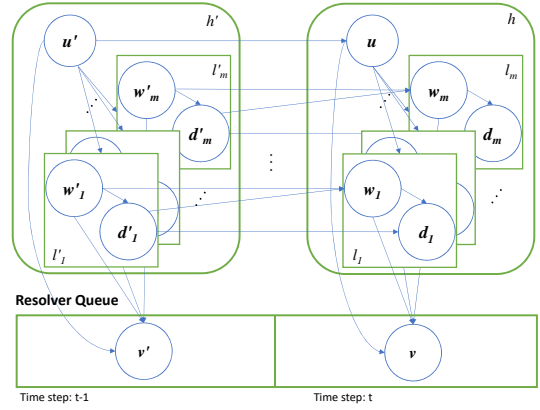}
		\caption{Illustration of the interleaving process. The random variable $u$ selects the user that generates the query for the time step $t$ and stalls the others. The selected user proceed according the user model, HsMM, and outputs the query $v$.}
		\label{fig:rq}
	\end{figure}

	We model the interleaving process as an Augmented Hidden Markov Model (AHMM), where the hidden states are augmented states, i.e., combination of variables \cite{minot2014separation}. 
	To make the equations more readable, we lump together the variables corresponding to each user and make the following lumped variable $L_k(t) = (W_{k}(t), D_{k}(t))$ and the hidden state of the HMM becomes $H(t) = (L_1(t), \dots, L_m(t), U(t))$ which is a $2m +1$ dimensional vector.
	Fig \ref{fig:rq} illustrates the interleaving process that leads to sequence generation.
	For simplicity, we assume $u(t-1) = u'$ and $u(t) = u$ which means that users $u'$ and $u$ are active at time steps $t-1$ and $t$ respectively. 
	At the time step $t$, user $u(t) = u \in [m]$ generates the request $v(t)$ which is the observed (visible) variable of the HMM.
	The request $v(t)$ is determined by the next request of the user in its HsMM model, i.e., $r_{u}(t)$.	
	Therefore, the emission probability of the AHMM is:
	{\small
	\be 
	\nr 
	\pr(V(t) = v(t) | H(t) = h(t)) = \pr_{u}(r_{u}(t) | w_{u}(t)) = \O_{u}(w_{u}(t), r_{u}(t)).
	\ee 
	}
	Now we derive the entries of the transition probability matrix of the AHMM:
	\be 
	\pr(H(t) | H(t-1)) = \alpha_{u'u} \prod_{k=1}^{m} \pr(l_k(t) | l_k(t-1), u),
	\ee 
	In the case of $k \neq u$ the user $k$ is not active, i.e., stalled. 
	Substituting the probability distributions from \eqref{eq:probs}, we get: 
	\be
	\pr(l_k(t) | l_k(t-1), u) = 
	\begin{cases}
		k \neq u & \delta(w, w')\delta(d, d') \\
		k = u & 
		\begin{cases}
			d = 0 & p_w(d) [\P_u]_{w'w} \\
			d > 0 & \delta(d, d-1)
		\end{cases}  		
	\end{cases}
	\ee 
	
	\begin{table}
		\centering
		\begin{tabular}{|c|l|}
			\hline
			Symbol & Explanation \\ 
			\hline 
			$L$ & The lumped random variable $L = (W, D)$. \\ \hline
			$H$ & The hyper-hidden state of the HMM $H = (L_1, \dots, L_m, U)$.\\ \hline
			$V$ & The visible state of the HMM which is the requested DN. \\ \hline 
		\end{tabular}
		\caption{Summary of the augmented variables. For each random variable the corresponding small letter represents a realization.}
	\end{table}

	\section{Deinterleaving Methods}
	In the deinterleaving problem, given $\{v(t)\}_{t=1}^T$ we are
        interested in inferring $\{u(t)\}_{t=1}^T$.  In other words,
        we want to find the users who initiated each request from the
        sequence generated by the interleaving process described in
        Section \ref{subsec:interle}.
	
	We present two candidate approaches for inference.  One is
        based on reducing the interleaving process to an AHMM as
        discussed in Section \ref{subsec:interle}.  This approach has
        been used for deinterleaving of Markov chains with small
        number of chains (users) and state space
        \cite{minot2014separation}.  Next, we propose to deinterleave
        using 	an LSTM model which have recently been shown to 
        perform well in many time-series analysis tasks\cite{chung2014empirical,Hochreiter}.
	
	\subsection{Inference on Augmented HMM}
        We can model the whole interleaving process as an AHMM
        and use learning techniques (like
        EM) to learn its parameters and use Viterbi inference to
        determine the most probable hidden (augmented) states $h(t)$
        from which we can extract the most probable user $u(t)$.  The
        main difficulty of applying this framework is that the state
        space of hidden variable, Figure \ref{fig:hsmm}, is very
        large.  More specifically, there are $m (nq)^m$ possible
        states of $h$ and as we increase number of webpages $n$ or
        users $m$ the state space grows exponentially.  The huge state
        space, makes the inference and learning very hard and as we
        show in Section \ref{subsec:viterbi} for synthetic
        experiments, even when the model parameters are known,
        deinterleaving performs (using Viterbi coding) poorly.
	
	\subsection{Inference using an LSTM}
	\label{subsec:lstm}
        RNNs \cite{lecun2015deep} are popular for modeling time
        series data. Given the input $v_t$ and hidden state $h_{t-1}$,
        the RNN computes the next hidden state representation $h_t$
        and output $u_t$ using the following recurrent relationships
	\begin{align}
	\label{eq:rnn}
	h_{t} &= f(W_v v_{t} + W_h h_{t - 1} + b)\\
	u_t &= W_u h_t
	\end{align}
	where $W_v$, $W_h$, $W_u$ and $b$ are the network parameters, and $f()$ is some non-linear function. 
        An example of $f$ could be a sigmoid $f(z) = \sigma(z) = 1/(1+\exp(-z))$ or rectified linear unit $f(z) = \max(0,z)$.

	For our specific problem of deinterleaving, an RNN can 
        by posed as a multi-class classification problem, where
        the input is the observed webpage and the output will be the
        identified user who requested that webpage. Specifically, each
        data instantiation consists of a sequence of user-request
        pairs, i.e., $(u(t), v(t))$.  This represents who was the user
        at a given time $t$ and what request was produced by that
        user. Both the user and the request are represented by an
        integer. The RNN is unrolled for the entire length of one
        sequence. The users and request integers are converted to
        one-hot encoding to enable learning. Thus if there are $b$
        possible web pages, the requests become $b$-dimensional
        vectors and for $m$ users, it becomes an $m$-dimensional
        vector. The request vectors are fed as input to the RNN model,
        while the output is the corresponding user at each
        time-step. The RNN’s $m$-dimensional output is passed through
        a softmax layer to convert it into probabilities and the user
        with higher probability is compared against the ground
        truth. Performance is measured in terms of accurately
        identifying the user at each instant.

	A common variant of RNN is LSTM	\cite{Hochreiter} which we use in our experiments. 
	We randomly initialize network parameters $W_v, W_h, W_u$ and
        apply stochastic gradient descent (SGD) (for RNNs, it is also
        referred to as a Backpropagation-through-time (BPTT)
        algorithm). In particular, we use a variation of the standard
        SGD called Adam \cite{kingma2014adam}, which allows for
        adaptive learning rates using the past gradients, similar to
        using momentum. This results in faster convergence compared to
        other adaptive algorithms.

	\section{Experiment}

	We start with a synthetic toy example and compare our LSTM algorithm with Viterbi inference as the baseline and then move to larger experiments.
	In all experiments, accuracy is measured as $\frac{1}{T}\sum_{t = 1}^{T} \indic(u_t = \hat{u}_t)$ where $u_t$ is the actual user generated 	query $t$ and $\hat{u}_t$ is the inferred user.

	\subsection{Viterbi vs. RNN}
	\label{subsec:viterbi}
	Here we generate synthetic resolver queue using the most complicated user model, i.e., HsMM of Figure \ref{fig:hsmm2} and report the Viterbi and LSTM methods performance. 
	To reduce the computational burden for the Viterbi algorithm, we restrict ourselves to 2 pages, 2 users, and 2 possible requests per page. 
	To make the setup even simpler, user $i$ browse only page $i$ and page $i$ picks from two possible requests at random using Beta(3+$\epsilon$,1+$\delta$) where $\epsilon$ and $\delta$ are independent and uniform over $[0,1]$.
	Viterbi is tested on the same sequences of size thousand. 
	Results are averaged over 5 realizations of the synthetic data. 
	With this setup the size of the hidden state space of the AHMM built from the HsMM user model is 32 and the number of observations is 2. 
	The users shares vector is $\aalpha = (.4, .6)$.
	LSTM is trained, validated and tested with sequences of size 6, 3, 1 thousands requests, respectively.
	
	Table \ref{tab:viterbi} summarizes the result:
	\begin{table}
		\scriptsize  
		\centering
		\begin{tabular}{|c|c|c|}
			\hline
			{\bf Method} & {\bf Viterbi} & {\bf LSTM} \\ 
			\hline  			
			Mean Accuracy	& 0.51	&  0.92  \\ \hline 
			Std of Accuracy	& 0.02	&  .17   \\ \hline 
		\end{tabular}
		\caption{Comparing accuracy of Viterbi coding and LSTM methods for the toy example. Results are averaged over 5 realization of the synthetic data. The baseline accuracy based on the proportion of users $\aalpha = (.4, .6)$ is $.6$.}
		\label{tab:viterbi}
	\end{table}
    Interestingly, LSTM outperforms Viterbi by a large margin. 
	Note that we perform Viterbi assuming that HMM parameters are given and not learned from data using algorithms like Baum--Welch, and even with this setup Viterbi performs poorly, worst that the baseline. 
	Perhaps, the poor performance of Viterbi compared with LSTM can be explained by the linear nature of Viterbi coding and the intrinsic power of LSTM in learning non-linear temporal relations. 
	
	\subsection{Synthetic Experiment}
	\label{subsec:synthetic}
	\begin{table}
		\centering
		\begin{tabular}{|c|l|}
			\hline
			Parameter & Value \\ 
			\hline  
			$m$ & 2 users \\ \hline
			$n$ & 20 pages \\ \hline
			$q$ & Maximum of 5 request per page\\ \hline 
			\hline 
			$\aalpha$ & $(0.4, 0.6)$ \\ \hline 
			$\A$ & Diagonal dominated row stochastic random matrix$^*$. \\ \hline 
			$\P_u$ & A random $20 \times 20$ matrix$^*$ \\ \hline 			
			$p_w(d)$ & $\text{Uniform}(1,5)$ \\ \hline 
			$\O_u$ & A random $20 \times 20$ matrix$^*$ \\ \hline 
		\end{tabular}
		\caption{Summary of the experimental setup for the synthetic experiment \ref{subsec:synthetic}. $^*$More on the random matrix generation in the text.}
		\label{tab:toy}
	\end{table}

	Owing to the poor performance of AHMM approach from now on we focus on LSTM method of Section \ref{subsec:lstm}.
	We report the results of seven synthetic experiments only for LSTM which is trained, validated and tested with sequences of size 60, 30, 10 thousands requests, respectively.
	
	We test the results for 7 different scenarios, in all of them we want to deinterleave a sequence generated by two users but the parameters in each experiment is set up differently.
	Table \ref{tab:toy} specifies the shared parameter setup. 
	Specific user transition and emission matrices are set for different scenarios which are explained in Section \ref{subsub:sparsity}.
	Note that in our experiments we report results on two set of synthetic data set, where in one we have a users shares vector $\aalpha$ determining the share of each user from the queue's requests.
	In the other more general data generating scheme, we assume that the users transition matrix $\A$ governs the turn in request submission.
	Different distributions for $\aalpha$ and $\A$ are discussed in Section \ref{subsub:dist}.

	{\bf Sparsity Patterns of Matrices:}
	\label{subsub:sparsity}
	For each user $u$ we have two matrices $\P_u$ and $\O_u$ which are randomly generated. 
	The generation process assumes that each row of both matrices is sparse, which is a reasonable assumption. 
	Each user view and surf a limited number of pages and on each page the possible requests are from a small subset of the all available pages. 
	The supports of $\P_i$s and $\O_i$s can overlap or be disjoint and this combination generates the different setups of our experiments.  
	After selecting a support we generate a discrete distribution over that support, which will be discussed in Section \ref{subsub:dist}.
	
	In the following the outer-list determines the different strategies for generating $\P_u$s and the inner-list elaborates the method of building $\O_u$s. 
	Each row of $\O_u$s has $a$ non-zero elements (randomly selected) and the distribution is uniform.
	We call $\O_1 \neq \O_2$ and $\O_1 = \O_2$ schemes, personalized and shared outputs respectively.
	
\begin{figure}
		\centering
		\begin{subfigure}[b]{0.15\textwidth}
			\includegraphics[width=\textwidth]{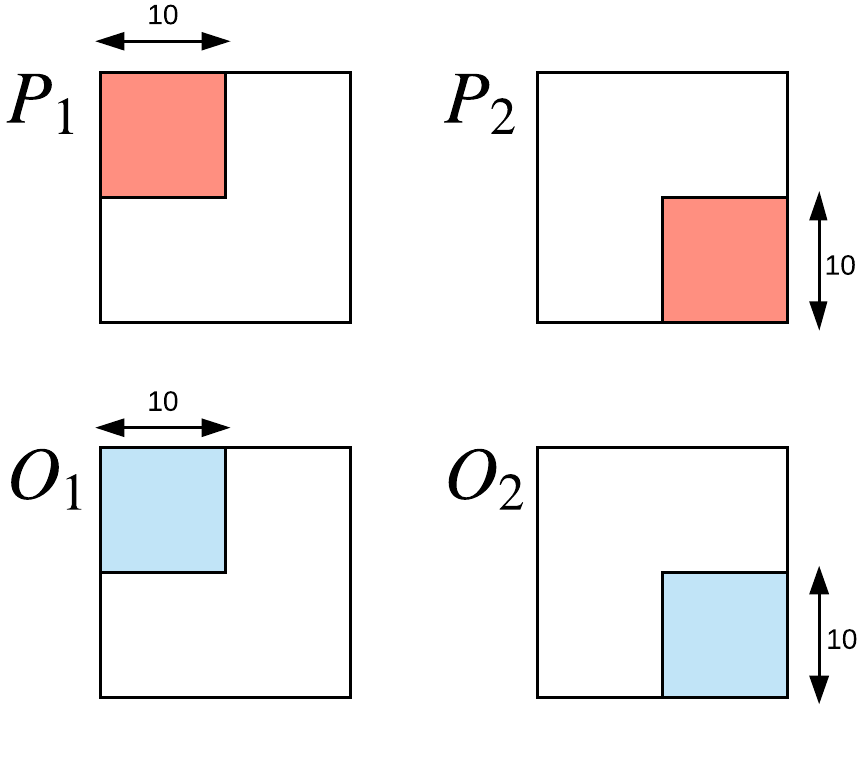}
			\caption{Case 1}
			\label{fig:case1}
		\end{subfigure}
		~ 
		\begin{subfigure}[b]{0.15\textwidth}
			\includegraphics[width=\textwidth]{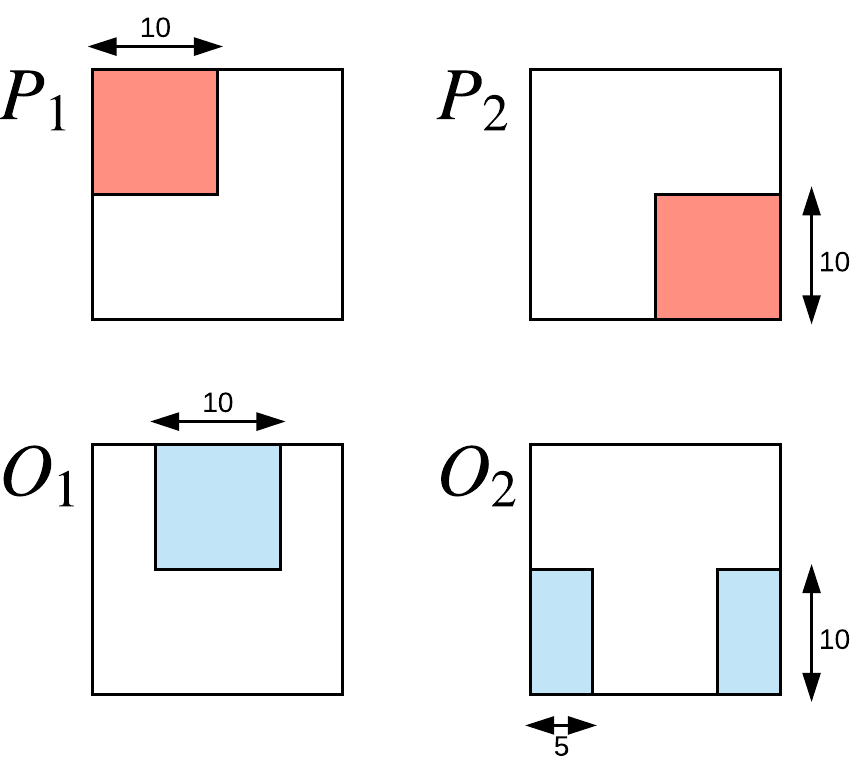}
			\caption{Case 2}
			\label{fig:case2}
		\end{subfigure}
		~ 
		\begin{subfigure}[b]{0.15\textwidth}
			\includegraphics[width=\textwidth]{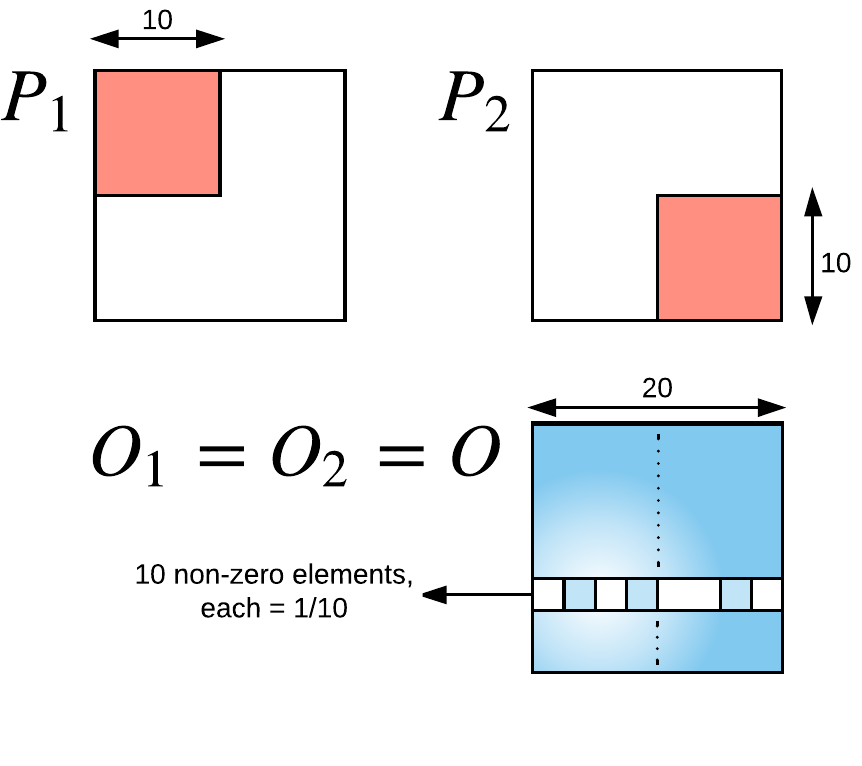}
			\caption{Case 3}
			\label{fig:case3}
		\end{subfigure}
		\caption{Illustration of the disjoint surfing categories for $a = 10$ and $b = 20$.}\label{fig:dsurf}
	\end{figure}

\begin{figure}
		\centering
		\begin{subfigure}[b]{0.2\textwidth}
			\includegraphics[width=\textwidth]{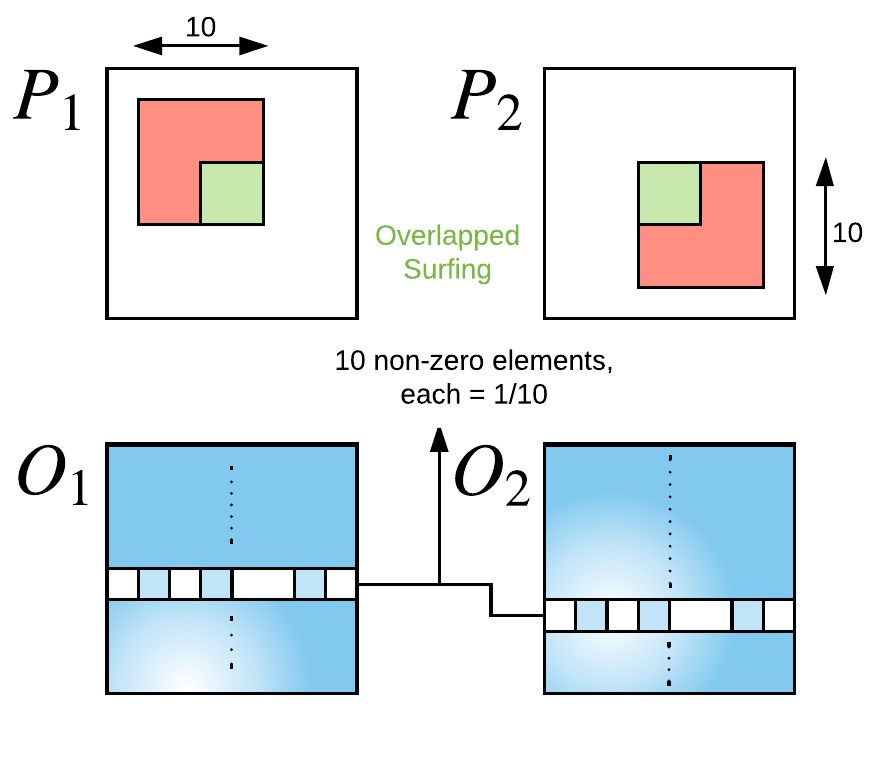}
			\caption{Case 4}
			\label{fig:case4}
		\end{subfigure}
		~ 
		\begin{subfigure}[b]{0.2\textwidth}
			\includegraphics[width=\textwidth]{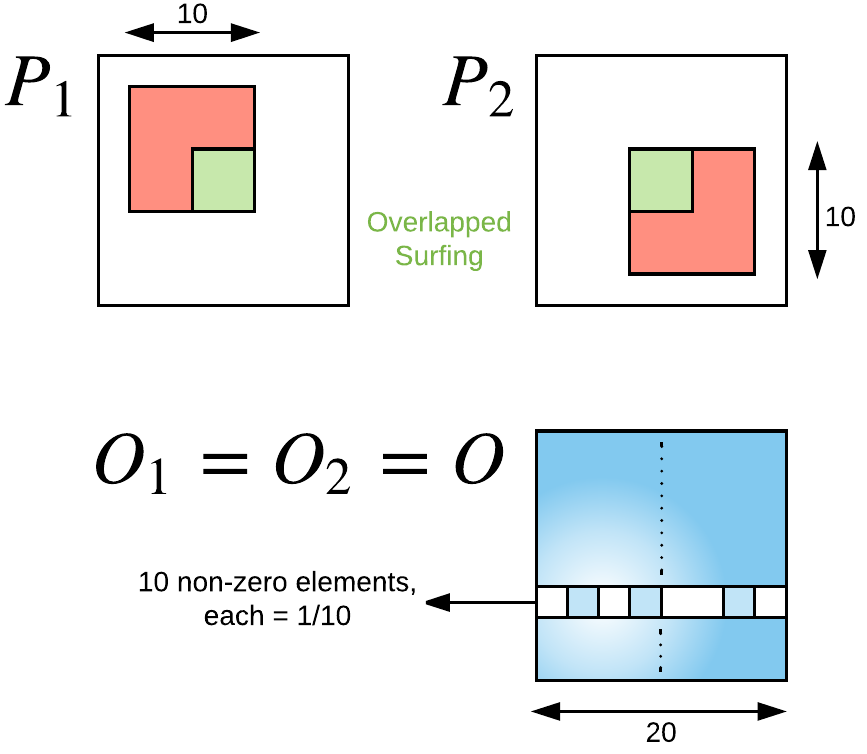}
			\caption{Case 5}
			\label{fig:case5}
		\end{subfigure}
		\caption{Illustration of the overlapped surfing without auxiliary block for $a = 10$ and $b = 20$.}\label{fig:osurf}
	\end{figure}

\begin{figure}
		\centering
		\begin{subfigure}[b]{0.2\textwidth}
			\includegraphics[width=\textwidth]{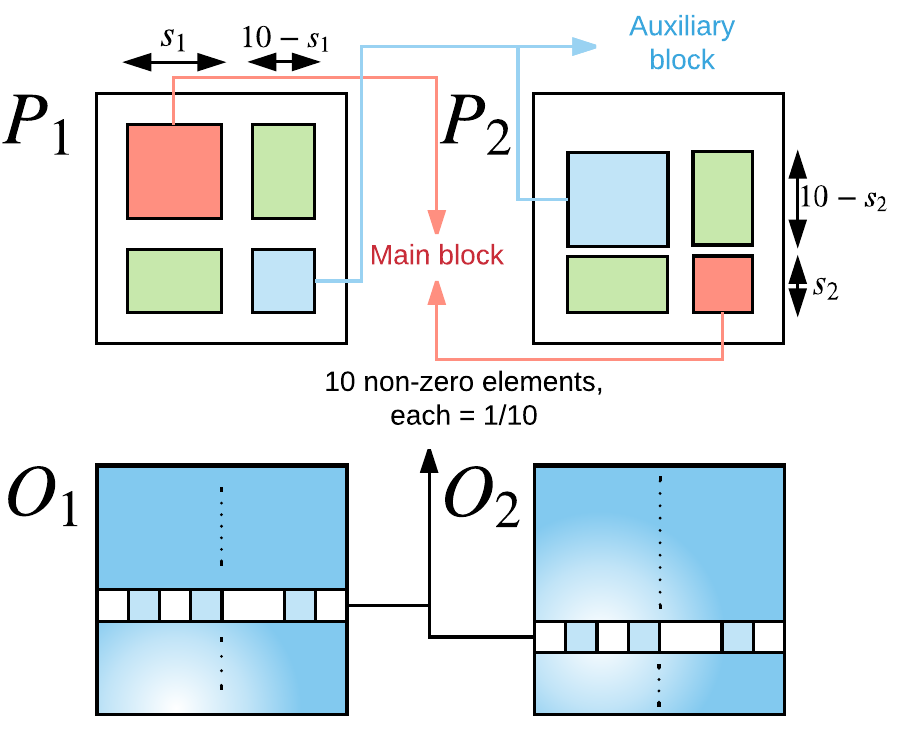}
			\caption{Case 6}
			\label{fig:case6}
		\end{subfigure}
		~ 
		\begin{subfigure}[b]{0.2\textwidth}
			\includegraphics[width=\textwidth]{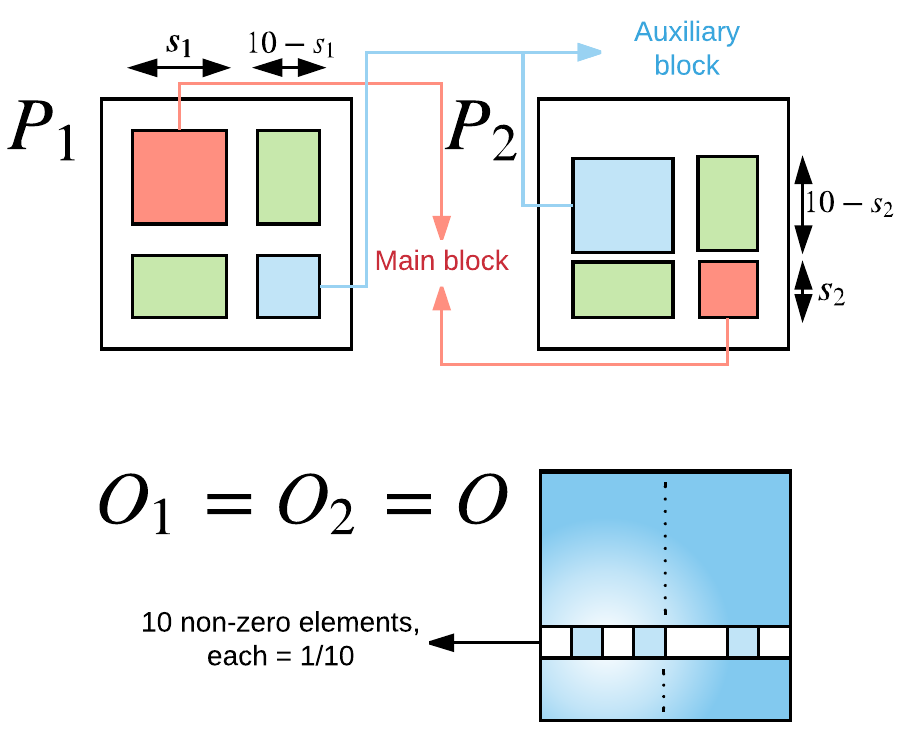}
			\caption{Case 7}
			\label{fig:case7}
		\end{subfigure}
		\caption{Illustration of the overlapped surfing with auxiliary block for $a = 10$ and $b = 20$.}\label{fig:osurfau}
	\end{figure}

	\begin{itemize}
		\item {\bf Disjoint webpage surfing:}  In this scenario, users surf disjoint parts of the web, say user 1 surf inside a group of first $a$ pages and user 2 surf the remaining $n - a$ pages, Fig \ref{fig:dsurf}.
		\begin{itemize}
			\item[] {\bf Case 1)} \emph{Disjoint personalized outputs - same grouping as webpages:} $\O_u$ and $\P_u$ have similar sparsity patterns, Fig \ref{fig:case1}. 
			\item[] {\bf Case 2)} \emph{Disjoint personalized outputs:} $\O_u$ and $\P_u$ do not have similar sparsity patterns, but support of $\O_1$ and $\O_2$ are disjoint, Fig \ref{fig:case2}. 
			\item[] {\bf Case 3)} \emph{Shared output:} Fig \ref{fig:case3}.
		\end{itemize}
		\item {\bf Overlapped webpage surfing with fixed block size: }
		Each user selects its surfing support of size $a$ at random. 
		Supports may overlap, Fig \ref{fig:osurf}. 
		\begin{enumerate}
			\item[] {\bf Case 4)} \emph{Personalized outputs:} Fig \ref{fig:case4}. 
			\item[] {\bf Case 5)} \emph{Shared output:} Fig \ref{fig:case5}.
		\end{enumerate}
		\item {\bf Overlapped webpage surfing with variable block size and interaction between blocks: }
		Each user selects $s = \text{Uniform}(1, a)$ pages at random as its main support (higher probability of surfing in these $s$ pages), and $a-s$ pages again at random as its auxiliary support (pages that user seldom visits), Fig \ref{fig:osurfau}.
		\begin{enumerate}
			\item[] {\bf Case 6)} \emph{Personalized outputs:} Fig \ref{fig:case6}. 
			\item[] {\bf Case 7)} \emph{Shared output:} Fig \ref{fig:case7}.
		\end{enumerate}
	\end{itemize}
	
	{\bf Probability Distributions:}
	\label{subsub:dist}
	Here we explain different distributions used in our synthetic generator:
	\begin{itemize}
		\item {Users shares vector $\aalpha$:} We fix $\aalpha$ to $(.4, .6)$.
		
		\item {Rows of user transition matrix $\A(u)$:} This is a diagonal dominant matrix, meaning that if user $u$ has submitted the current request $v(t)$ it is more probable that he submit the next request. 
		In this way, we capture the fact that because of the episodic nature of the request submission, close-by queries are more probable to come from a same user. 
		This has been exploited in the previous literature \cite{gao2013empirical}. Note that when instead of matrix $\A$ we only consider the vector $\aalpha$ we may not capture this realistic property of the data.
		For the toy example, we set $\alpha_{ii} = .5 + \frac{1}{2} \text{Uniform}(0,1)$ and $\forall i \neq j: \alpha_{ij} \propto (1 - \alpha_{ii}) \text{Uniform}(0,1)$
		
		\item {Rows of output matrix $\O_u(w)$:} As mentioned before, each row $\O_u(w)$ has $a$ non-zero elements with each with probability $1/a$.
		
		\item {Rows of page transition matrix $\P_u(w)$:} In a nutshell, we discretize a continuous Beta distribution with different parameters for each block and normalize the final vector.
		The distribution that we use for the (main) support is Beta(3+$\epsilon$,1+$\delta$) where $\epsilon$ and $\delta$ are random numbers from $[-1, 1]$. 
		For the distribution on the auxiliary support of the cases 6 and 7 above, we use Beta(2+$\epsilon$,2+$\delta$).
	\end{itemize}

	{\bf Discussion:}
	Table \ref{tab:as} shows the error of our method for all 7 cases of Section \ref{subsub:sparsity} for $a = 10$. 
	
	\begin{table}
		\scriptsize  
		\centering
		\begin{tabular}{|c|c|c|c|c|c|c|c|c|}
			\hline
			\multicolumn{2}{|c|}{\bf \footnotesize Cases} & {\bf \footnotesize  1} & {\bf \footnotesize  2} & {\bf \footnotesize  3} & {\bf \footnotesize  4} & {\bf \footnotesize  5} & {\bf \footnotesize 6} & {\bf \footnotesize  7}  \\ 
			\hline  
			\multirow{2}{*}{$\aalpha$} & {\footnotesize  Mean} 	&  1 	& 1		&  .63 &  .70 & .62 &  .74& .65 \\ \cline{2-9}
			                           & {\footnotesize  Std} 	&  0 	& 0 	&  .02 &  .02 & .02 &  .05& .04 \\ \hline 
			\hline  
			\multirow{2}{*}{$A$}       & {\footnotesize  Mean} 	&  1 	& 1		&  .77 & .69 &  .67 & .82 & .78 \\ \cline{2-9}
			                           & {\footnotesize  Std} 	&  0 	& 0		&  .10 & .09 &  .08 & .09 & .16 \\ \hline 
		\end{tabular}
		\caption{Deinterleaving accuracy of LSTM for different cases of the synthetic example when user transitions are determine by either of $\aalpha = (.4, .6)$ or random diagonally dominant $\A$. The baseline for $\aalpha$ and $\A$ experiments are $.6$ and $.5$ respectively.}
		\label{tab:as}
	\end{table}
	
	Each row is the average result for five instantiation of the model parameters $\O_u$ and $\P_u$. 
	The error of each instantiation (each row) is an average of 100 experiments. 
	Note that case 1 and 2 are trivial cases when both $\P$s and $\O$s are disjoint and LSTM perfectly dis-interleave. 
	Interestingly, performance in case 3 is much worse than cases 1 and 2, which confirms that in our model having disjoint output matrices is more important than disjoint surfing pattern. 
	Intuitively, this makes sense because the final request comes from the output matrices and if we have personalized outputs the deinterleaving should be easier. 
	Interestingly, beyond the trivial cases 1 and 2, case 6 has the best accuracy, probably because of personalized outputs and more complicated $\P_u$ for each user (composed of main and auxiliary block) makes the whole problem more separable.

    \section{Conclusion}

This paper describes our foray into the application of advanced 
deep-learning techniques to the problem of deinterleaving DNS-based
time-series sequences.  To this end, we developed an HsMM-based model
of user request generation and an AHMM-based model of the interleaving
process at the resolver queue.  We then evaluated the efficacy of two
different inference strategies for deinterleaving on a synthetic
dataset.  Our results suggest that LSTM-based strategies significantly
outperform traditional AHMM-based models.  In future work, we plan to
extend this analysis on signficantly larger datasets to the specific
problem of malware domain group extraction.

	\section*{Acknowledgments}
	The work was supported in part by NSF grants CNS 1314560, IIS-1447566, IIS-1447574, IIS-1422557, CCF 1451986, and IIS-1563950. SG and VY acknowledge partial support from NSF Grant CNS-1314956 and CNS-1514503.
	
\bibliographystyle{plain}
\bibliography{main}

\begin{thebibliography}{10}

\bibitem{bahdanau2014neural}
Dzmitry Bahdanau, Kyunghyun Cho, and Yoshua Bengio.
\newblock Neural machine translation by jointly learning to align and
  translate.
\newblock {\em arXiv preprint arXiv:1409.0473}, 2014.

\bibitem{batu2004inferring}
Tugkan Batu, Sudipto Guha, and Sampath Kannan.
\newblock Inferring mixtures of markov chains.
\newblock In {\em COLT}, volume 2004, pages 186--199. Springer, 2004.

\bibitem{burge1997prediction}
Chris Burge and Samuel Karlin.
\newblock Prediction of complete gene structures in human genomic dna.
\newblock {\em Journal of molecular biology}, 268(1):78--94, 1997.

\bibitem{burge1998finding}
Christopher~B Burge and Samuel Karlin.
\newblock Finding the genes in genomic dna.
\newblock {\em Current opinion in structural biology}, 8(3):346--354, 1998.

\bibitem{chung2014empirical}
Junyoung Chung, Caglar Gulcehre, KyungHyun Cho, and Yoshua Bengio.
\newblock Empirical evaluation of gated recurrent neural networks on sequence
  modeling.
\newblock {\em arXiv preprint arXiv:1412.3555}, 2014.

\bibitem{gao2013empirical}
Hongyu Gao, Vinod Yegneswaran, Yan Chen, Phillip Porras, Shalini Ghosh, Jian
  Jiang, and Haixin Duan.
\newblock An empirical reexamination of global dns behavior.
\newblock {\em ACM SIGCOMM Computer Communication Review}, 43(4):267--278,
  2013.

\bibitem{NIPS2008_3449}
A.~Graves and J.~Schmidhuber.
\newblock Offline handwriting recognition with multidimensional recurrent
  neural networks.
\newblock In {\em NIPS}, pages 545--552. 2009.

\bibitem{Hochreiter}
S.~Hochreiter and J.~Schmidhuber.
\newblock Long short-term memory.
\newblock {\em Neural Comput.}, 9(8):1735--1780, November 1997.

\bibitem{kingma2014adam}
Diederik Kingma and Jimmy Ba.
\newblock Adam: A method for stochastic optimization.
\newblock {\em arXiv preprint arXiv:1412.6980}, 2014.

\bibitem{landwehr2008modeling}
Niels Landwehr.
\newblock Modeling interleaved hidden processes.
\newblock In {\em Proceedings of the 25th international conference on Machine
  learning}, pages 520--527. ACM, 2008.

\bibitem{lecun2015deep}
Yann LeCun, Yoshua Bengio, and Geoffrey Hinton.
\newblock Deep learning.
\newblock {\em Nature}, 521(7553):436--444, 2015.

\bibitem{minot2014separation}
Ariana Minot and Yue~M Lu.
\newblock Separation of interleaved markov chains.
\newblock In {\em Signals, Systems and Computers, 2014 48th Asilomar Conference
  on}, pages 1757--1761. IEEE, 2014.

\bibitem{seroussi2009deinterleaving}
Gadiel Seroussi, Wojciech Szpankowski, and Marcelo~J Weinberger.
\newblock Deinterleaving markov processes via penalized ml.
\newblock In {\em Information Theory, 2009. ISIT 2009. IEEE International
  Symposium on}, pages 1739--1743. IEEE, 2009.

\bibitem{seroussi2012deinterleaving}
Gadiel Seroussi, Wojciech Szpankowski, and Marcelo~J Weinberger.
\newblock Deinterleaving finite memory processes via penalized maximum
  likelihood.
\newblock {\em IEEE Transactions on Information Theory}, 58(12):7094--7109,
  2012.

\bibitem{sophos}
SophosLabs.
\newblock Looking ahead: Sophoslabs 2017 malware forecast.
\newblock 2017.

\bibitem{sutskever2014sequence}
I.~Sutskever, O.~Vinyals, and Q.~V. Le.
\newblock Sequence to sequence learning with neural networks.
\newblock In {\em NIPS}, pages 3104--3112, 2014.

\bibitem{symantec}
Symantec.
\newblock 2017 internet security threat report.
\newblock 2017.

\end{thebibliography}

\end{document}